\documentclass{article}

     \usepackage[nonatbib, preprint]{neurips_2021}

\usepackage[utf8]{inputenc} % allow utf-8 input
\usepackage[T1]{fontenc}    % use 8-bit T1 fonts
\usepackage{hyperref}       % hyperlinks
\usepackage{url}            % simple URL typesetting
\usepackage{booktabs}       % professional-quality tables
\usepackage{amsfonts}       % blackboard math symbols
\usepackage{nicefrac}       % compact symbols for 1/2, etc.
\usepackage{microtype}      % microtypography
\usepackage{graphicx}
\usepackage{amsmath,amssymb, amsthm}
\usepackage{mathrsfs}
\usepackage{natbib}
\usepackage[ruled]{algorithm2e} % For algorithms
\usepackage[dvipsnames]{xcolor}
\usepackage{natbib}
\setcitestyle{authoryear,open={(},close={)}} %Citation-related commands

\usepackage{enumitem}
\DeclareMathOperator*{\amax}{arg\,max}
\def\B{{B }}
\def\R{\mathbb{R}} 
\def\E{\mathbb{E}}
\def\f{objective function }

\def\x{design vector }

\def\xnospace{design vector}
\def\xs{design vectors }
\def\X{design space }
\def\y{observation }
\def\ys{observations }
\def\yc{performance }

\def\xr{recommended design }
\def\xrnospace{recommended design}

\def\acq{\text{cKG}}
\def\acqspace{\text{cKG }}
\def\Cov{\text{Cov}\,}
\def\Fn{\mathscr{D}_{f}} % curly F^n
\def\Cn{\mathscr{D}_{c}}%
\def\mun{expected objective performance }
\title{Bayesian Optimisation for Constrained Problems}

\author{
  Juan Ungredda\\
  University of Warwick\\
  Coventry, CV4 7AL, UK \\
  \texttt{j.ungredda@warwick.ac.uk} \\
   \And
   Juergen Branke \\
  University of Warwick\\
  Coventry, CV4 7AL, UK \\
  \texttt{juergen.branke@wbs.ac.uk} \\
}
\newtheoremstyle{mystyle}%                % Name
  {}%                                     % Space above
  {}%                                     % Space below
  {\itshape}%                                     % Body font
  {}%                                     % Indent amount
  {\bfseries}%                            % Theorem head font
  {.}%                                    % Punctuation after theorem head
  { }%                                    % Space after theorem head, ' ', or \newline
  {}%                                     % Theorem head spec (can be left empty, meaning `normal')
\theoremstyle{mystyle}

\newtheorem{thm}{Theorem}
\newtheorem{lm}{Lemma}

\begin{document}

\maketitle

\begin{abstract}
Many real-world optimisation problems such as hyperparameter tuning in machine learning or simulation-based optimisation can be formulated as expensive-to-evaluate black-box functions. A popular approach to tackle such problems is Bayesian optimisation (BO), which builds a response surface model based on the data collected so far, and uses the  mean and uncertainty predicted by the model to decide what information to collect next.
In this paper, we propose a novel variant of the well-known Knowledge Gradient acquisition function that allows it to handle constraints. We empirically compare the new algorithm with four other state-of-the-art constrained Bayesian optimisation algorithms and demonstrate its superior performance.
We also prove theoretical convergence in the infinite budget limit.

%however, many results in unconstrained functions show a superior performance when the impact of a sample \x into future iterations is taken into account. However, extending look-ahead BO algorithms to constraints have been largely unexplored and represents a challenging task. We propose a novel acquisition function  that takes into account the one-step-look ahead impact on both the objective and constraints. We also prove theoretical convergence in the infinite budget limit and show that the proposed approach outperforms available benchmarks in different synthetic experiments.

\end{abstract}

\section{Introduction}

Expensive black-box constrained optimisation problems appear in many fields where the possible number of evaluations is limited. Examples include hyperparameter tuning, where the objective is to minimise the validation error of a machine learning algorithm \citep{Hernandez2016}, the optimisation of the control policy of a robot
under performance and safety constraints \citep{Berkenkamp2016BayesianOW}, or engineering design optimisation \citep{Forrester08}.

For such applications, Bayesian optimisation (BO) has shown to be a powerful and efficient tool. After collecting some initial data, BO constructs a surrogate model, usually a Gaussian process (GP). Then it iteratively uses  an acquisition function to decide what data should be collected next. The Gaussian process model is updated with the new sample information and the process is repeated until the available budget of evaluations has been consumed. Most BO approaches assume unconstrained or box-constrained problems. 
%A few variants for constrained problems exist, however, they ignore the impact of the new sample on the future objective . As a consequence, this may affect the location of the estimated optimum of the model, specially when the optimum is located at the feasibility boundary. This motivates the development of a look-ahead method that includes constraints, which brings the following contributions,

In this paper, we make the following contributions.
\begin{enumerate}
    \item We develop a new variant of the Knowledge Gradient acquisition function, called \emph{constrained Knowledge Gradient} (cKG), capable of handling constraints.
    \item We show how cKG can be efficiently computed.
    \item We prove that cKG converges to the optimal solution in the limit.
    \item We apply our proposed approach to a variety of test problems and show that cKG outperforms other available BO approaches for constrained problems.
\end{enumerate}

We start with an overview of related work in Section~\ref{sec:LitRew}, followed by a formal definition of the problem in Section~\ref{sec:ProbDef}. Section~\ref{sec:Alg} explains the statistical models, shows the suggested sampling procedure, outlines its theoretical properties and computation. We perform numerical experiments in Section \ref{sec:Experiments}. Finally, the paper concludes with a summary and some suggestions for future work in Section \ref{sec:Conclusions}.

\section{Literature Review}\label{sec:LitRew}

Bayesian optimisation (BO) has gained wide popularity, especially for problems involving expensive black-box functions, for a comprehensive introduction see \cite{frazier2018tutorial} and \cite{deFreitas2016}. Although most work has  focused on unconstrained problems, some extensions to constrained optimisation problems exist.

Many of the approaches are based on the famous Expected Improvement (EI) acquisition function  \citep{Jones1998}. 
 \cite{schonlau1998} and \cite{Gardner2014} extended EI  to constrained EI (cEI) by computing the expected improvement of a point $x$ over the best feasible point and multiplying it by its probability of being feasible. This relies on the assumption that the objective function and constraints are independent, and that the decision maker is risk neutral. \cite{Bagheri17} proposed a modified combination of probability of feasibility with EI that makes it easier to find solutions on the feasibility boundary.

Several extensions to EI have also been proposed for noisy problems. \cite{Letham2017} extended Expected Improvement to noisy observations (NEI) and noisy constraints by iterating the expectation over possible posterior distributions. For noise-free observations, their approach reduces to the original cEI. Other methods rely on relaxing the constraints instead of modifying the infill criteria, \cite{Gramacy2016} proposed an augmented Lagrangian approach that includes constraints as penalties in the objective function. \cite{Picheny2016} refined the previous approach by introducing slack variables and achieve better performance on equality constraints. 

The Knowledge Gradient (KG) policy (\cite{Frazier211}) is another myopic acquisition function that aims to maximise the new predicted optimal performance after one new sample, and can be directly applied to either deterministic or noisy functions. \cite{chen2021new} recently proposed an extension of KG to constraints by multiplying any new sampling location by its probability of feasibility. These approaches consider noise in the observations and constraints but they only use the current feasibility information. In contrast, \cite{Lam2017} consider deterministic problems and proposed a lookahead approach for the value of feasibility information, selecting the next evaluation in order to maximise the long-term feasible increase of the objective function. This was formulated using dynamic programming  where each simulated step gives a reward following cEI. 
%The aforementioned methods either assume non-noisy observations or they do not take into account how the feasibility would change in following iterations.

Other acquisition functions not based on EI have also been considered to tackle constraints. \cite{Hernandez-Lobato2016}  extended Predictive Entropy Search \citep{Hernandez-Lobato2014} to constraints. This acquisition criterion involves computing the expected entropy reduction of the global solution to the constrained optimization problem.  \cite{Eriksson2019} extended Thompson sampling for constrained optimisation and also proposed a trust region to limit the search to locations close to the global optimum. \cite{picheny2014} proposed an optimisation strategy where the benefit of a new sample is measured by the reduction of the expected volume of the excursion set which provides a measure of uncertainty on the minimiser location where constraints can be incorporated in the formulation by a solution's probability of being feasible. However this can only be computed approximately using numerical integration. \cite{Antonio2019SequentialMB}  proposed a two-stage approach where the feasible region is estimated during the first stage by a support-vector classifier, then the second stage uses the estimated boundaries and maximises the objective function value using 
 the Upper Confidence Bound (UCB) as acquisition function.

\section{Problem Definition}\label{sec:ProbDef}
We want to find the optimizer $x^{*}$ of a black-box function $f: \mathbb{X} \rightarrow \mathbb{R}$ with constraints $c_{k}:\mathbb{X} \rightarrow \mathbb{R}$, i.e., 

\begin{align}
\begin{split}
x^{*} = \amax_{x \in \mathbb{X}} f(x)
\end{split}\\
\begin{split}
    \mbox{s.t.} \text{  }& c_{k}(x) \leq 0 \text{  }, k=1,\dots,K
\end{split}
\end{align}

The \f $f$ takes as arguments a \x $x \in \mathbb{X} \subset \mathbb{R}^{d}$ and returns an \y corrupted by  noise $y = f(x) + \epsilon$, where $\epsilon \sim N(0, \sigma^{2}_\epsilon)$, and a vector of %noiseless 
constraint values  $\mathbf{c} = [c_{1}(x), \dots, c_{K}(x)]$.\\

There is a total budget of $B$ samples that can be spent. After consuming the budget, a \xrnospace, $x_{r}$, is returned to the user and its quality is determined by the difference in \f 
%between a \xr $x_{r}$ and 
to the best solution $x^{*}$ given that $x_{r}$ is feasible, i.e. $x \in F=\{x|c_{k}(x)\leq 0\quad \forall k\in [1\ldots K] \}$. If $x_{r}$ is not feasible then there is a penalty $M$ for not having a feasible solution. Therefore, the quality of a solution may be measured as an Opportunity Cost (OC) to be minimised,

\begin{equation}
  OC(x_{r}) =
    \begin{cases}
      f(x^{*}) - f(x_{r}) & \text{if $x_{r} \in F$}\\
      f(x^{*}) - M & \text{otherwise}
    \end{cases}       
\end{equation}

Without loss of generality we assume a penalty $M=0$. However, $M$ may be set by using the minimum GP estimate of the \f in the \X \citep{Letham2017}.

\section{The cKG Algorithm}\label{sec:Alg}

\subsection{Statistical Model}

Let us denote all $n$ \xs sampled so far as $X = \{x_{i}\}_{i=1}^{n}$  and the training data from the collection of \f \ys, $\Fn=\{(x,y)\}_{i=1}^{n}$, and constraints,$\Cn=\{(x,\mathbf{c})\}_{i=1}^{n}$. We model the \f \ys as a Gaussian process (GP) which is fully specified by a mean function $\mu_{y}^n(x) = \mathbb{E}[y(x)|\Fn]$ and its covariance $\Cov[y(x),y(x')|\Fn]$,

\begin{align}
\begin{split}
\mathbb{E}[y(x)|\Fn] &= \mu_{y}^n(x) \\
&=\mu_{y}^0(x)-k_{y}^0(x,X)(k_{y}^0(X,X)+I\sigma^{2}_{\epsilon})^{-1}(Y-\mu_{y}^0(X))
\end{split}\\
\begin{split}
\Cov[y(x),y(x')|\Fn] &= k_{y}^n(x,x')\\
&= k_{y}^0(x,x')-k_{y}^0(x,X)(k_{y}^0(X^n,X^{n})+I\sigma^{2}_{\epsilon})^{-1}k_{y}^0(X,x')).
\end{split}
\end{align}

Similarly, each constraint is modelled as an independent GP over the training data $\Cn$ defined by a constraint mean function $\mu_{k}^n(x) = \mathbb{E}[c_{k}(x)|\Cn]$ and covariance $\Cov[c_{k}(x),c_{k}(x')|\Cn]$. The prior mean is typically set to zero and the kernel
allows the user to encode known properties such as smoothness
and periodicity. We use the popular squared exponential kernel that assumes
$f$ and $\mathbf{c}$ are smooth functions, i.e., nearby $x$  have similar outputs while widely separated points have unrelated outputs. 
%We estimate hyper-parameters by maximising the marginal likelihood. 
Further details can be found in \cite{Rasmussen06}.

\subsection{Recommended Solution}

% At the end of the algorithm we must recommend a final feasible \xnospace, $x_{r}$. Assuming a risk-neutral user, the utility of a solution is its expected quality if feasible, and zero if infeasible. Therefore, if the constraints and objective are independent, a recommended solution $x_{r}$ may be obtained by

At the end of the algorithm we must recommend a final feasible \xnospace, $x_{r}$. Assuming a risk-neutral user, the utility of a \x is the \mun, $\mu_{y}^n(x)$, if feasible, and zero $(M)$ if infeasible. Therefore, if the constraints and objective are independent, a recommended solution $x_{r}$ may be obtained by

%such that it maximises the \f $f$, as,

%\begin{align}
%\begin{split}
%x_{r} = \amax_{x \in \mathbb{X}}\mathbb{E}_{y, \mathbf{c}}[y(x) %\mathbb{I}_{x \in F}|\Fn, \Cn]
%\end{split}
%\end{align}

%where $\mathbb{I}_{x \in F}$ is a feasibility indicator. Although $f$ and $c_{k}$ are unknown, their statistical approximations, $\mu_{y}^{B}(x)$ and $\mu_{k}^B(x)$, are available. Assuming a risk-neutral user, the utility of a solution is its expected quality if feasible, and zero if infeasible. Therefore, if the constraints and objective are independent, a recommended solution $x_{r}$ may be obtained by

\begin{eqnarray}
x_{r} = \amax_{x \in \mathbb{X}}{\mu_{y}^{B}(x) \mathbb{P}[\mathbf{c}^{B}(x)\leq 0] },
\end{eqnarray}\label{equ:recom_sample}

where $\mathbb{P}[\mathbf{c}^{B}(x)\leq 0] $ is the probability of feasibility of a design $x$. Following \cite{Gardner2014}, we assume independent constraints, such that $ \mathbb{P}[\mathbf{c}^{B}(x)\leq 0| \Cn] = \prod_{k=1}^{K}\mathbb{P}[c^{B}_{k}(x)\leq 0| \Cn]$. Each term $\mathbb{P}[c_{k}(x)\leq 0| \Cn]$ can be evaluated by a univariate Gaussian cumulative distribution. In the remainder of this work we denote the probability of feasibility as $\text{PF}^{n}(x)$. 

% \begin{align}
% \begin{split}
%  \mathbb{P}[\mathbf{c}^{B}(x)\leq 0| \Cn] =\mathbb{P}[c^{B}_{1}(x) \leq 0, \dots, c^{B}_{K}(x) \leq 0 | \Cn] = \prod_{k=1}^{K}\mathbb{P}[c^{B}_{k}(x)\leq 0| \Cn],
% \end{split}\label{Eq:final_performance}
% \end{align}

%The quality of the recommended solution is the difference between its actual utility, and the utility of the best feasible solution, which is often called Opportunity Cost:
%\begin{eqnarray}
%OC(x_r)=\left\{
%begin{array}{rl}
%\max_{x\in F} y(x)- y(x_r) & x_r\in F\\
%\max_{x\in F} y(x) & \mbox{otherwise}
%\end{array}
%\right.
%\end{eqnarray}
% Finally, we aim to choose the designs $x_{1}, \dots, x_{B}$ that maximises the expected value of the estimated performance,

% $$
% x_{r} = arg\max_{x \in \mathbb{X}}\mathbb{E}[ {\mu_{y}^{B}(x) \mathbb{P}[\mathbf{c}^{B}(x)\leq 0}] \big| x^{1}=x_{1}, \dots, x^{B} =x_{B} ]
% $$

% where the expected value is taken over the uncertainty of evaluating $x_{1}, \dots, x_{B}$. 

\subsection{Acquisition Function}

We aim for an acquisition function  that quantifies the value of the objective function and constraint information we would gain from a given sampling decision.
%benefit and interaction between the \f and constraints given a sampling decision. However,
Note that obtaining feasibility information does not immediately translate to better \mun but rather more accurate feasibility information where more updated feasibility information may change our current beliefs about where $x_{r}$ is located. Therefore, to quantify the benefit of a design vector, we first find the \xr given by the sampled trained data $\Cn$ and $\Fn$ as,

\begin{align}
\begin{split}
x^{n}_{r} = \amax_{x \in \mathbb{X}}{\mu_{y}^{n}(x)\text{PF}^{n}(x)  }.
\end{split}\label{equ:recom_alg_sample}
\end{align}

A sensible compromise between the current step $n$ and the one-step lookahead estimated \yc is offered by augmenting the training data by the sampling decision $x^{n+1}$ with its respective constraint and objective observations as $\Cn \cup \{x^{n+1}, \mathbf{c}^{n+1}\}$ and $\Fn \cup \{x^{n+1}, y^{n+1}\}$. The difference in \yc between the current \xr and the new best \yc presents an acquisition function for a design $x$,

\begin{align}
\begin{split}
\acq(x) &= \mathbb{E}[\max_{x' \in \mathbb{X}}\big\{\mu_{y}^{n+1}(x') \text{PF}^{n+1}(x')  \big\} - \mu_{y}^{n+1}(x^{n}_{r}) \text{PF}^{n+1}(x^{n}_{r}) | x^{n+1} = x].
\end{split}\label{eq:formulation_1}
\end{align}

Eqn.~\ref{eq:formulation_1} is positive for all the \X and $\mu_{y}^{n+1}(x^{n}_{r})$ may be marginalised over $y^{n+1}$, such that 

\begin{align}
\begin{split}
\acq(x) &= \mathbb{E}[\max_{x' \in \mathbb{X}}\big\{\mu_{y}^{n+1}(x') \text{PF}^{n+1}(x')  \big\} - \mu_{y}^{n}(x^{n}_{r}) \text{PF}^{n+1}(x^{n}_{r}) | x^{n+1} = x].
\end{split}\label{eq:formulation}
\end{align}

This acquisition function quantifies the benefit of a \x and takes into account the change in the current \yc value when more feasibility information is available. Also, when constraints are not considered, the formulation reduces to standard KG \citep{Frazier211}. Theoretical guarantees can be proven for $\acq$. This policy ensures in a finite search space $\mathbb{X}$, with an infinite sampling budget all points will be sampled infinitely often which ensures learning the true expected \y (Theorem~\ref{thm:infinite_state_visitation}). Also, in the limit, cKG will find the true optimal solution $x^*$ (Theorem~\ref{thm:reverese_xstar}).

\subsection{Efficient Acquisition Function Computation}

Obtaining a closed-form expression for cKG is not possible but as we show below, it can still be  computed efficiently. \cite{pearce2020practical} proposed an efficient one-step-lookahead computation that  consists of obtaining high value points for different realisations of the posterior GP mean given a sample \x. Those discrete \xs can then be used as a discretisation in the \X for which  a closed-form solution exists. This approach is both computationally efficient and scalable with the number of \x dimensions, thus we adapt this method to our constrained problem. 

We first convert $\mu_{y}^{n+1}(x)$ to quantities that can be computed in the current step $n$ through the parametrisation trick \citep{Frazier211} as $ \mu_{y}^{n+1}(x)  = \mu_{y}^{n}(x) + \tilde{\sigma_{y}}(x,x^{n+1})Z_{y}$ where $Z_{y} \sim N(0,1)$. The deterministic function $\tilde{\sigma}^{n}(x,x^{n+1})$ represents the standard deviation of $\mu_{y}^{n+1}(x)$ parametrised by $x^{n+1}$ and given by, $\tilde{\sigma}^{n}(x,x^{n+1}) = \frac{k^{n}(x,x^{n+1})}{\sqrt{k^{n}(x^{n+1}, x^{n+1}) + \sigma^{2}_{\epsilon}}}$.

% \begin{align}
% \begin{split}
% \tilde{\sigma}^{n}(x,x^{n+1}) = \frac{k^{n}(x,x^{n+1})}{\sqrt{k^{n}(x^{n+1}, x^{n+1}) + \sigma^{2}_{\epsilon}}}.
% \end{split}
% \end{align}

Similarly, we may apply the parametrisation trick to the posterior means and variances of the constraints, i.e, $\mu_{k}^{n+1}(x)= \mu_{k}^{n}(x) + \tilde{\sigma_{k}}(x,x^{n+1})Z_{k}$ and $k^{n+1}_{k}(x,x) = k^{n}_{k}(x,x) - \tilde{\sigma_{k}}^{2}(x,x^{n+1})$, where $Z_{k} \sim N(0,1)$ for $k=1, \dots, K$. Now, the probability of feasibility is also parametrised by $x^{n+1}$ and all the stochasticity is determined by $\mathbf{Z}_{c}=[Z_{1}, \dots, Z_{K}]$. By plugging these parametrisations into  Eqn. \ref{eq:formulation}, we change our initial problem to variables that can be estimated in the current step where the stochasticity is given by standard normally distributed random variables for both constraints and the objective,

% \begin{align}
% \begin{split}
% \mu_{k}^{n+1}(x)= \mu_{k}^{n}(x) + \tilde{\sigma_{k}}(x,x^{n+1})Z_{k}
% \end{split}\\
% \begin{split}
% k^{n+1}_{k}(x,x) = k^{n}_{k}(x,x) - \tilde{\sigma_{k}}^{2}(x,x^{n+1})
% \end{split}
% \end{align}
\begin{align}
\begin{split}
\acq(x) &= \mathbb{E} \Bigg[ \overbrace{\max_{x' \in \mathbb{X}}\big\{\big [ \mu_{y}^{n}(x') + \tilde{\sigma}_{y}(x',x^{n+1})Z_{y} \big ] \text{PF}^{n+1}(x';x^{n+1},\mathbf{Z}_{c}) }^\text{Inner Optimisation $n+1$} \big\} \nonumber
\end{split}\\[3ex]
\begin{split}
& - \mu_{y}^{n}(x_{r}) \text{PF}^{n+1}(x^{n}_{r};x^{n+1},\mathbf{Z}_{c})| x^{n+1} = x  \Bigg].
\end{split}\label{eq:formulation_parametrised}
\end{align}

To solve the above expectation we first find $x^{n}_{r}$ according to Eqn. \ref{equ:recom_alg_sample} using a continuous numerical optimiser. Then, given a design $x^{n+1}$, we generate $n_{y}$ values from $Z_{y}$ and $n_{c}$ values from $\mathbf{Z_{c}}$ where the inner optimisation problems in Eqn.~\ref{eq:formulation_parametrised} is solved by a continuous numerical optimiser for all $n_{z} = n_{c} * n_{y}$ values. Each solution found by the optimiser, $x_{j}^{*}$, represents a peak location, and together they determine a discretisation $X_{d} = \{x_{1}^{*}, \dots, x_{n_{z}}^{*}\}$. Finally, Eqn.~\ref{eq:formulation_parametrised} can be solved in closed-form where now the inner optimisation problems are computed over the discrete set $X_{d}$. Conditioned on $\mathbf{Z}_{c}$, the above expectation can be seen as marginalising the standard discrete KG \citep{Frazier211} over the constraint uncertainty where $\mu_{y}^{n}(x)$ and $\tilde{\sigma_{y}}(x,x^{n+1})$ is penalised by the (deterministic) function $\text{PF}^{n+1}(x;x^{n+1},\mathbf{Z}_{c})$. Therefore, if we denote each standard KG computation as $\text{KG}_{d}(x^{n+1}=x; \mathbf{Z}_c)$ we may compute the overall expectation by a Monte-Carlo approximation,

\begin{align}
\begin{split}
\acq(x) &= \frac{1}{n_{c}} \sum_{m=1}^{n_{c}} \text{KG}_{d}(x^{n+1}=x; \mathbf{Z}^{m}_c).
\end{split}\label{eq:formulation_hybrid}
\end{align}

\begin{figure}
	\centering
    	\begin{tabular}{ccc}
    	\includegraphics[height=3.cm]{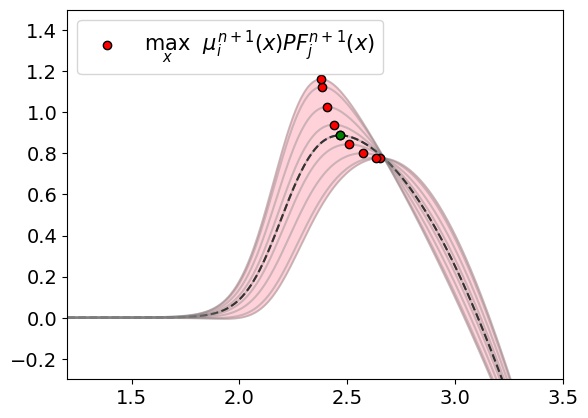}&				\includegraphics[height=3.cm]{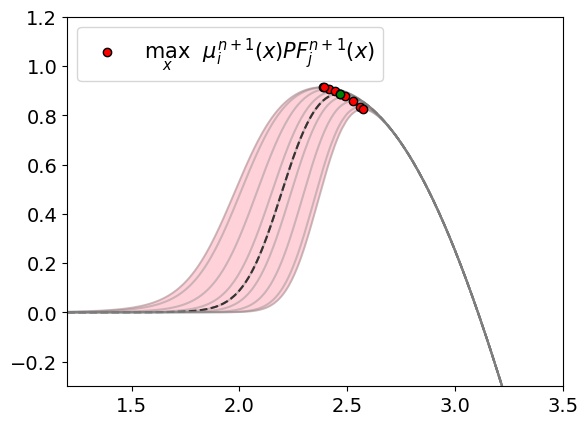}&
    	\includegraphics[height=3.4cm]{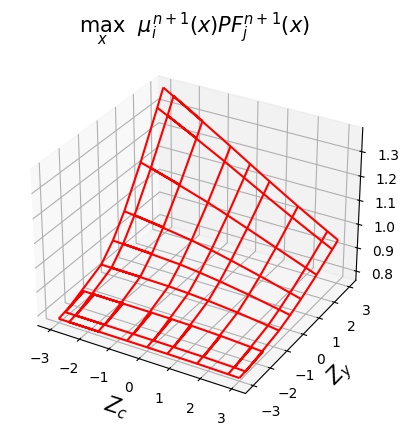}\\
    	(a)&(b)&(c)\\
    	\end{tabular}\caption{(a) Given $\mathbf{Z_{c}}=0$, current GP mean (dotted grey) and maximum (green dot) where changing $Z_{y}$ produces a different realisation and a new maximum (red dots). (b) Given $Z_{y}=0$, different values of $\mathbf{Z}_{c}$ produces a new maximum according to the probability of feasibility. (c) shows the surface of the maximum posterior over the discrete set for all combinations of $\mathbf{Z}_{c}$ and $Z_{y}$.}
	    \label{fig:computations}
\end{figure}

Fig.~\ref{fig:computations} shows the influence of a sample $x^{n+1}$ on computing the expectation at $n+1$ in Eqn.~\ref{eq:formulation_parametrised}. More specifically, if we fix $\mathbf{Z}_{c}$, Fig.~\ref{fig:computations} (a) shows how the current GP mean (dotted grey) and maximum (green dot) could change according to $Z_{y}$ where each different realisation presents a new maximum (red dots). However, if we fix $Z_{y}$, Fig.~\ref{fig:computations} (b) shows how the maximum of the GP mean may change according to the probability of feasibility. Fig.~\ref{fig:computations} (c) shows the surface of the maximum locations for all combinations of $\mathbf{Z}_{c}$ and $Z_{y}$.

\subsection{Overall Algorithm}

\begin{figure}
\centering
\includegraphics[height=0.5cm]{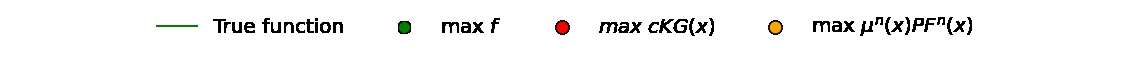}\\
	\centering
    	\begin{tabular}{ccc}
    	\includegraphics[height=2.8cm]{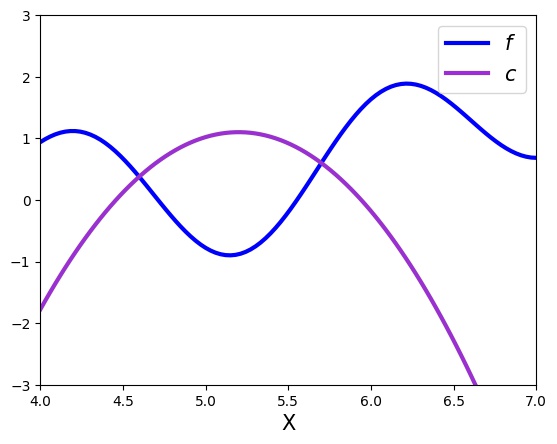}&	
    	\includegraphics[height=2.8cm]{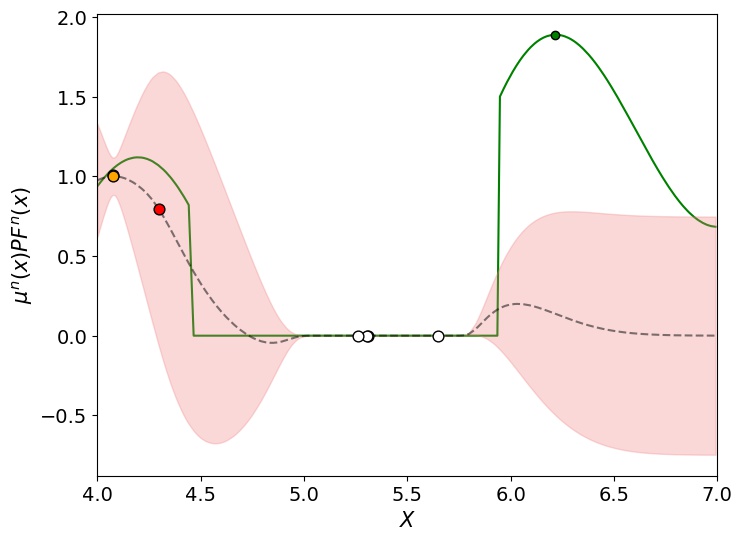}&	
    	\includegraphics[height=2.8cm]{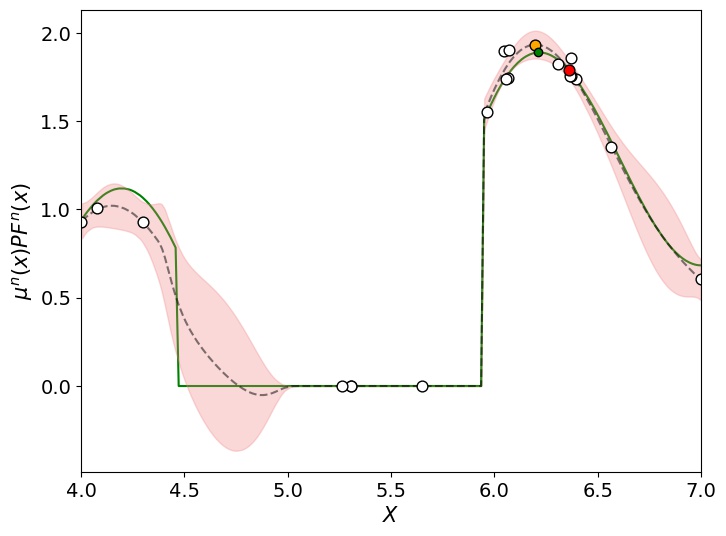}\\
    	(a)&(c)&(e)\\
    	
    	\includegraphics[height=2.8cm]{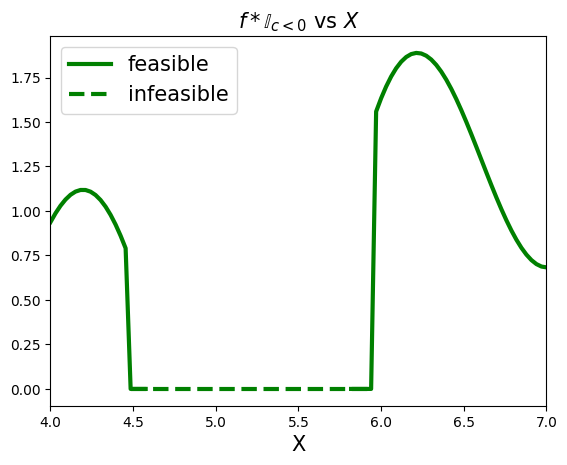}&	
    	\includegraphics[height=2.8cm]{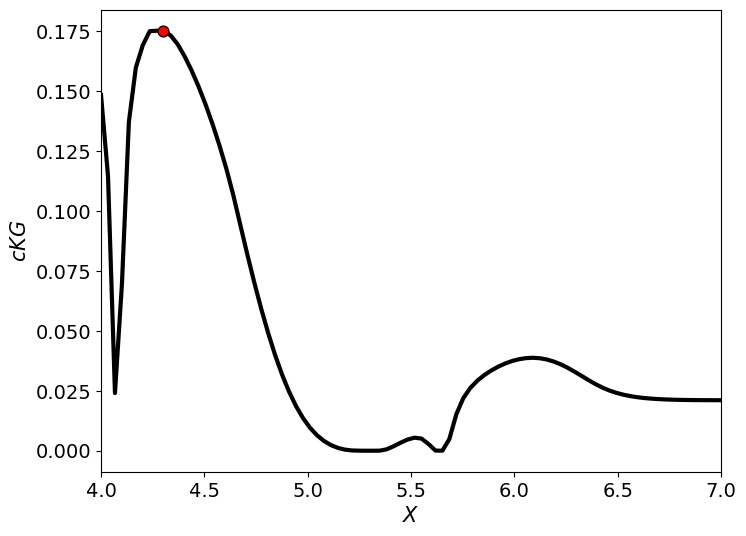}&				
    	\includegraphics[height=2.8cm]{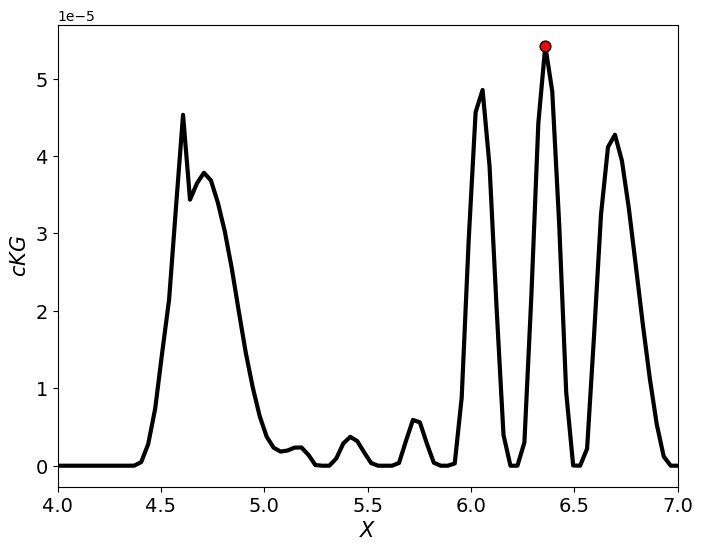}\\
    	(b)&(d)&(f)\\
    	\end{tabular}\caption{(a) \f and constraint where constraint values less than zero are feasible. (b) Feasible and infeasible regions with its corresponding values. (c) Initial design allocation where a model is built using a GP for the \f penalised by the probability of feasibility using a GP for the constraints. (d) and (f) show the next sample decision (red dot) according to cKG using the fitted models. (e) shows the samples taken during the entire optimisation run (white dots) with the \xr (orange dot) coinciding with the true best \x (green dot).}
	\label{fig:iterations}
\end{figure}

Fig.~\ref{fig:iterations} shows iterations of cKG. Fig.~\ref{fig:iterations} (a) shows an \f (blue) and a constraint (purple) with negative constraint values representing feasible solutions. The aim is to find the best feasible solution at $x^*=6.25$, see also Fig.~\ref{fig:iterations} (b). Then, GPs are built based on initial samples, and (c) shows the posterior utility (dotted line, posterior $y$ times probability of feasibility). Then, the next \x is obtained my maximising cKG. Finally, after the budget of $B$ samples has been allocated sequentially, a final recommendation $x_{r}$ is selected according to Eqn. \ref{equ:recom_alg_sample} where $x_{r}$ (orange dot) ends up being very close to the true best $x^*$ (green dot). Notice that cKG aims at improving the maximal posterior mean, not the quality at the sampled solution, and thus often tends to sample the  neighborhood of $x^*$ instead of the actual best \x location. 

% On the other hand, a risk-averse user may not be happy with a recommended solution that hasn't been evaluated, and thus that is not guaranteed to be feasible. Therefore, at the last iteration, $B$, we use noisy Expected Improvement (nEI) such that it increases the chances of sampling \xs closer to $x^{*}$ before giving a final recommendation. See Appendix \ref{App:algorithms_background} for a brief description of nEI.

cKG is outlined in Algorithm \ref{alg:cKG}. On Line 1, the algorithm begins by fitting a Gaussian process model to the initial training data $\Fn$ and $\Cn$ obtained using a Latin hypercube (LHS) ‘space-filling’ experimental design.  After initialisation, the algorithm continues in an optimisation loop until the budget $B$ has been consumed. In each iteration, we sample a new \x $x^{n+1}$ according to cKG, as defined in Algorithm \ref{alg:cKG_computation} (Line 2).  The \x that maximises cKG determines the sample $(x,y)^{n+1}$ and $(x,\mathbf{c})^{n+1}$. The point is added to the training data $\Fn$ and $\Cn$ and each Gaussian process model is updated (Line 5).  Finally, cKG recommends a \x according to Eqn.~\ref{equ:recom_sample} (Line 7). More implementation details may be found in Appendix \ref{App:Implementation Details}.

% Just before the budget is depleted, at step $B$, a last sample is taken according to nEI (Line 7).

\begin{algorithm}
	\caption{cKG computation.
	}\label{alg:cKG_computation}

	\KwIn{Sample $x^{n+1}$, size of Monte-Carlo discretisations $n_{c}$ and $n_{y}$}
	\vspace{1.2mm}
	0. Initialise discretisation $X^{0}_{d}=\{\}$ and set $n_{z}= n_{c} n_{y}$\\
    \vspace{1.2mm}
	1. Compute $x^{n}_{r} = \amax_{x \in \mathbb{X}}{\mu_{y}^{n}(x)\text{PF}^{n}(x)}$\\
    \vspace{1.2mm}
	2.  \textbf{for} j\textbf{ in }[1, \dots,  $n_{z}$] \textbf{:}\\
	\vspace{1.0mm}
    3.\hspace{5mm}Generate $Z^{j}_{y}, Z^{j}_{1}, \dots, Z^{j}_{K} \sim N(0,1)$\\
    \vspace{0.5mm}
    4.\hspace{5mm}Compute $x_{j}^{*} = \max_{x \in X_{d}} \big\{\big [ \mu_{y}^{n}(x) + \tilde{\sigma_{y}}(x,x^{n+1})Z^{j}_{y} \big ] \text{PF}^{n+1}(x;x^{n+1},\mathbf{Z}^{j}_{c}) \big\}$  \\
    \vspace{0.5mm}
    5.\hspace{5mm}Update discretisation $X^{j}_{d} =X^{j-1}_{d} \cup \{x_{j}^{*}\} $\\
    \vspace{0.5mm}
    6. \textbf{for} m\textbf{ in }[1, \dots,  $n_{c}$] \textbf{:}\\
    \vspace{0.5mm}
    7.\hspace{5mm}Compute $\text{KG}_{d}(x^{n+1}=x; \mathbf{Z}^{m}_c)$ using $X_{d}$\\
    \vspace{0.5mm}
    8. Compute Monte-Carlo estimation $\frac{1}{n_{c}} \sum_{m=1}^{n_{c}} \text{KG}_{d}(x^{n+1}; \mathbf{Z}^{m}_c)$\\
    9. \textbf{Return:} $\acq(x^{n+1})$ 
\end{algorithm}

\begin{algorithm}
	\caption{cKG Overall Algorithm. The algorithm starts with an initialization phase
	to collect preliminary data, then, proceeds to a sequential phase.
	}\label{alg:cKG}

	\KwIn{
	black-box function $f:X \to \R$, constraints $c_{k}: X \to \R$,  size of Monte-Carlo $n_{c}$ and $n_{y}$}
	\vspace{1.2mm}
    0. Collect initial simulation data, $\Fn$,$\Cn$, and fit an independent Gaussian process for each constraint and the black-box function. \\
    \vspace{1.2mm}
	1. \textbf{While} \text{$b$ < \B} \textbf{do:}\\
	\vspace{1.0mm}
    2. \hspace{5mm}Compute $x^{n+1} = \arg \max_{x \in X }\text{cKG}(x, n_{z}, M)$.\\
    \vspace{0.5mm}
    3. \hspace{5mm}\text{Update $\Fn$, } with sample $\{(x,y)^{n+1}\}$\\
    4. \hspace{5mm}\text{Update $\Cn$, } with sample $\{(x,\mathbf{c})^{n+1}\}$\\
    \vspace{0.5mm}
    5. \hspace{5mm}\text{Fit a Gaussian process to $\Fn$ and $\Cn$} \\
    \vspace{0.5mm}
    6. \hspace{5mm}\text{}Update budget consumed, $b \gets b + 1$\\
    \vspace{0.5mm}
    7. \textbf{Return:} Recommend solution, $x_{r} = \arg\max_{x \in \mathbb{X}}\{\mu_{y}^{B}(x) \text{PF}^{B}(x)\}$ 
\end{algorithm}

    % \vspace{0.5mm}
    % 7. Compute $x^{B-1} = \arg \max_{x \in X }\text{nEI}(x)$.\\

\section{Experiments}\label{sec:Experiments}

In this section, we compare cKG against a variety of well-known acquisition functions that can deal with constraints, including: constrained Expected Improvement (cEI) by \cite{Gardner2014}, expected improvement to noisy observations (NEI) by \cite{Letham2017}, Predictive Entropy Search with constraints (PESC) by \cite{Hernandez-Lobato2016}, Thompson sampling for constrained optimisation (cTS) by \cite{Eriksson2019} and a recently proposed constrained KG algorithm \citep{chen2021new} which we call penalised KG (pKG) to distinguish it from our proposed formulation (further details on the benchmark algorithms can be found in Appendix \ref{App:algorithms_background}). 
%Note that this approach only performs a one-step-lookahead on the \f and it works by penalising standard KG with the probability of feasibility. 

We used  implementations of cEI and NEI available in BoTorch \citep{balandat2020botorch}. For PESC, only the Spearmint optimisation package provided an available implementation of the algorithm that included constraints. The remaining algorithms have been re-implemented from scratch and can be accessed through github\footnote{The code for this paper is available at https://github.com/xxx/xxx (will be published after acceptance)}.

For all test problems, we fit an independent Gaussian process for each constraint $c_{k}$ and the black-box objective function $y$ with an initial design of size $10$ for the synthetic test functions and $20$ for the MNIST experiment, both chosen by Latin Hypercube Sampling. Also, for each Gaussian process, an RBF kernel is assumed with hyperparameters tuned by maximum likelihood, including the noise $\sigma^{2}_{\epsilon}$ in case of noisy problems.

\subsection{Synthetic Tests}

We test the algorithms on three different constrained synthetic problems: Mistery function, Test function 2, and New Branin from \cite{Sasenaphdthesis}. Each function was tested with and without a noise level $\sigma_{\epsilon}^{2}=1$ for the objective value (the constraint values were assumed to be deterministic). All synthetic test results were averaged over 30 replications and generated using a computing cluster. Further details of each function can be found in  Appendix \ref{App:synthetic_tests}.

 Fig.~\ref{fig:synthetic_results} shows the results of these experiments. Fig.~\ref{fig:synthetic_results} (\textbf{top row}) depicts a contour plot of each \f over its feasible area. The location of the optimum is highlighted by a green cross. Mistery and New Branin both have a single non-linear constraint whereas the infeasible area in  Test function~2 is the result of a combination of 3 different constraints.  Fig.~\ref{fig:synthetic_results} (\textbf{middle row}) shows the convergence of the opportunity cost over the number of iterations, for the case without noise.  As can be seen, cKG outperforms all benchmark approaches on the Branin  and Mistery function, with cEI second best. On Test Function~2, pKG converges to the same quality as cKG, with the other methods performing much worse. Overall cKG is the only method that consistently yields  superior performance across all three test problems.   Fig.~\ref{fig:synthetic_results} (\textbf{bottom row}) shows the performance when observations are corrupted by noise. Since cEI was designed for deterministic problems, it was replaced by the more general NEI for the noisy problems. 
 %However, cEI can be viewed as a special case of nEI when noise is not present. 
 Not surprisingly, in all  cases the performance of the different considered approaches deteriorated compared to the deterministic setting. The difference between  cKG and the other methods is even more apparent, with no method coming close to cKG's performance on any of the benchmarks. This shows that cKG is particularly capable of handling noisy constrained optimisation problems.
 
%  Fig.~\ref{fig:synthetic_results} shows the results of these experiments. Fig.~\ref{fig:synthetic_results} (\textbf{top row}) depicts a contour plot of each \f over its feasible area. The location of the optimum is highlighted by a green cross. Mistery and New Branin both have a single non-linear constraint whereas the infeasible area in  Test function 2 is a result of 3 different constraints.  Fig.~\ref{fig:synthetic_results} (\textbf{middle row}) shows the convergence of the opportunity cost over the number of iterations, for the case without noise.  As can be seen, cKG outperforms all benchmark approaches on the Branin  and Mistery function, with cEI second best. On Test Function 2, pKG converges to the same quality as cKG, with the other methods performing much worse. Overall cKG is the only method that yields consistently superior performance across all three test problems.   Fig.~\ref{fig:synthetic_results} (\textbf{bottom row}) shows the performance when observations are corrupted by noise. Since cEI was designed for deterministic problems, it was replaced by the more general NEI for the noisy problems. 
%  %However, cEI can be viewed as a special case of nEI when noise is not present. 
%  Not surprisingly, in all  cases the performance of the different considered approaches deteriorated compared to the deterministic setting. The difference between  cKG and the other methods is even more apparent, with no method coming close to cKG's performance on any of the benchmarks. This shows that cKG is particularly capable of handling noisy constrained optimisation problems.

\begin{figure}[h]
	\centering
    	\begin{tabular}{ccc}
    	Branin & Mistery & Test Function 2\\
    	
    	\hspace{2mm}\includegraphics[height=3.15cm]{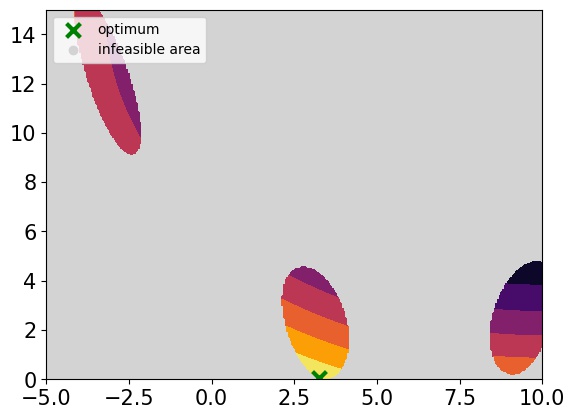}&
    	\hspace{2mm}\includegraphics[height=3.15cm]{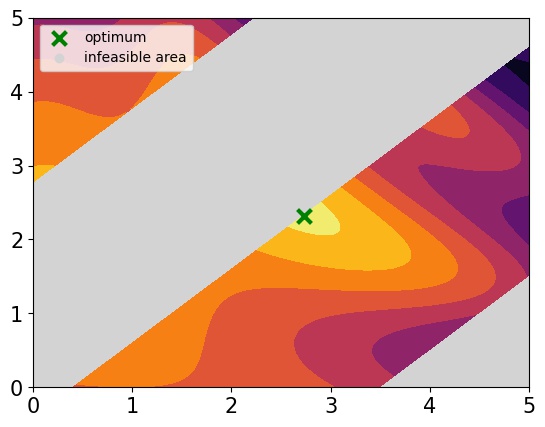}&
    	\hspace{2mm}\includegraphics[height=3.15cm]{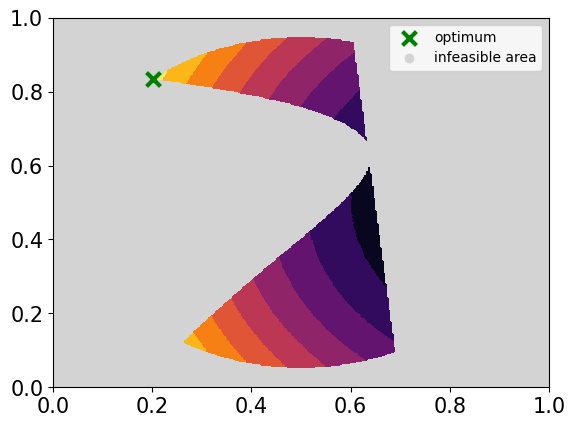} \\
    	
    	\includegraphics[height=2.8cm]{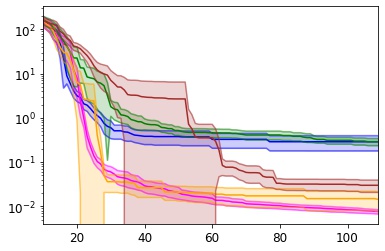}&				\includegraphics[height=2.8cm]{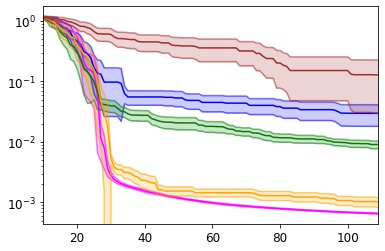}&
    	\includegraphics[height=2.8cm]{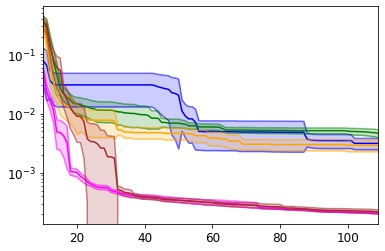}\\
    	
    	\includegraphics[height=2.8cm]{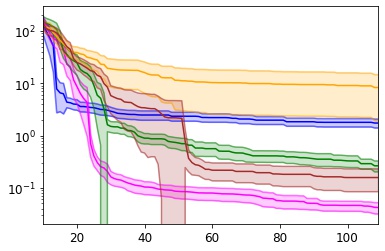}&				\includegraphics[height=2.8cm]{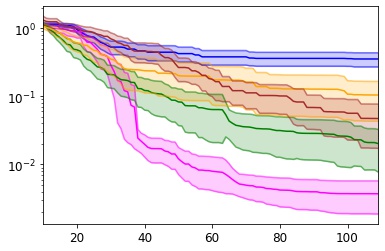}&
    	\includegraphics[height=2.8cm]{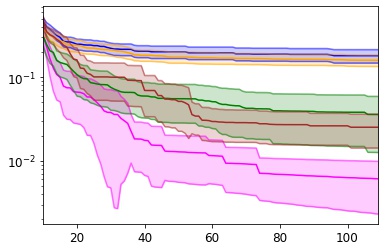}\\
    	\end{tabular}
    	\includegraphics[height=1.0cm, trim={0cm 3cm 0cm 3cm}, clip]{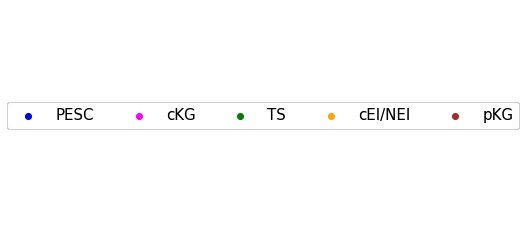}
        \caption{(\textbf{top row}) shows the feasible and infeasible regions of the considered synthetic functions. (\textbf{middle row}) and (\textbf{bottom row}) show the mean and 95\% CI for the OC over iterations for $\sigma_{\epsilon}^{2} =0$ and $\sigma_{\epsilon}^{2} =1$, respectively.}
	\label{fig:synthetic_results}
\end{figure}

\subsection{Tuning a Fast Fully Connected Neural Network}

For this experiment we aim to tune the hyperparameters of a fully connected neural network subject to a limit on the prediction time of 1 ms. The \X consists of 9 dimensions comprising the optimiser parameters and the number of neurons on each level, details of the neural network architecture may be found in Appendix \ref{MNIST Hyperparameter Experiment}. The prediction time is computed as the average time of 3000 predictions for minibatches of size 250. The network is trained on the MNIST digit classification task using tensorflow and the objective to be minimised is the classification error rate on a validation set.  Each \xr is evaluated 20 times to compute a "ground-truth" validation error. All results were averaged over 20 replications and generated using a 20-core Intel(R) Xeon(R) Gold 6230 processor. 

\begin{figure}[h]
	\centering
    	\begin{tabular}{cc}
    	
    	\includegraphics[height=3.2cm]{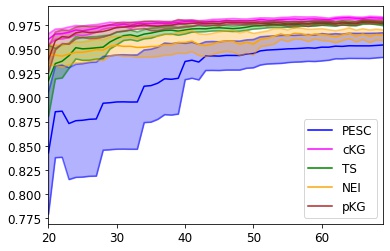}&				\includegraphics[height=3.2cm]{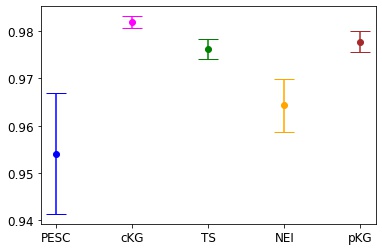}\\
    	
    	(a)&(b)\\
    	\end{tabular}
        \caption{(a) Mean and 95\% CI for the "ground-truth" validation accuracy over iterations. (b) Mean and 95\% CI for the "ground-truth score" after 50 iterations. }
        %All results were averaged over 20 replications}
	\label{fig:NN_results}
\end{figure}

Fig. \ref{fig:NN_results} shows that cKG yields the highest validation accuracy compared to the other considered benchmark methods. TS and pKG also perform well, which is consistent with the synthetic experiments.

\section{Conclusion}\label{sec:Conclusions}

For the problem of constrained Bayesian optimisation, we proposed a new variant of the well-known Knowledge Gradient acquisition function, constrained Knowledge Gradient (cKG), that is capable of handling constraints and noise. We show that cKG can be efficiently computed by adapting an approach proposed in \cite{pearce2020practical} which is a hybrid  between discretisation and Monte-Carlo approximation that allows to leverage the benefits of fast computations of the discrete \X and the scalability of continuous Monte-Carlo sampling. We prove that the algorithm will find the true optimum in the limit. Finally, we empirically demonstrate the effectiveness of the proposed approach on several test problems. cKG consistently and significantly outperformed all benchmark algorithms on all test problems, with a particularly large improvement under noisy problem settings.

%show the benefit of computing a one-step-lookahead in both the \f and the constraints when a sample \x is considered. Then, we propose a novel and efficient to compute acquisition function (cKG). These computations rely in a hybrid approach between discrete and Monte-Carlo approximations that allows to leverage the benefits of fast computations of the discrete \X and the scalability of continuous Monte-Carlo sampling. Finally, we empirically demonstrate the effectiveness of the proposed approach on several test problems.

Despite the excellent results, the study has some limitations that should be addressed in future work. First, while cKG should also work well with stochastic constraints, all the test problems considered here had a deterministic constraint function. Second, we have set the reward for an infeasible solution ($M$) to zero. A further study on the influence of this value may be interesting. Third, as most Bayesian optimisation algorithms, we assume the noise in the quality measure to be homoscedastic. Perhaps ideas from Stochastic Kriging can be used to relax this. Finally, we assume that an evaluation of a solution returns simultaneously its quality as well as its constraint value. In practice, it may be possible to evaluate quality and feasibility independently. 

%There are various directions of future research. We should test the algorithm also in case of noisy constraints, and extend the proposed approach to more than one-lookahead step,  batch sampling or the case where problems and constraints can be sampled independently. Also, further study on the influence of the penalisation constant $M$ would be warranted.

%how to properly penalise sampling decisions during the optimisation. 

% Other settings are straighforward to consider given the above formulation where the \f can be queried independently to the constraints. In this case, our approach may efficiently quantify the impact of sampling each "data source" and determine the source that presents greater benefit in future performance. 
%Strong assumptions in our paper
%\begin{itemize}
%    \item Problems with relatively low dimensionality. High %dimensionality sol: techniques as Turbo may be used.
%    \item Noise must be homocedastic. sol: Learn a landscape for the %noise.
%    \item We only test on deterministic constraints (?).
%\end{itemize}

\section*{Acknowledgements} Removed for double blind review

\bibliography{Bibliography}

\begin{thebibliography}{28}
\providecommand{\natexlab}[1]{#1}
\providecommand{\url}[1]{\texttt{#1}}
\expandafter\ifx\csname urlstyle\endcsname\relax
  \providecommand{\doi}[1]{doi: #1}\else
  \providecommand{\doi}{doi: \begingroup \urlstyle{rm}\Url}\fi

\bibitem[Antonio(2019)]{Antonio2019SequentialMB}
C.~Antonio.
\newblock Sequential model based optimization of partially defined functions
  under unknown constraints.
\newblock \emph{Journal of Global Optimization}, pages 1--23, 2019.

\bibitem[Bagheri et~al.(2017)Bagheri, Konen, Allmendinger, Branke, Deb,
  Fieldsend, Quagliarella, and Sndhya]{Bagheri17}
S.~Bagheri, W.~Konen, R.~Allmendinger, J.~Branke, K.~Deb, J.~Fieldsend,
  D.~Quagliarella, and K.~Sndhya.
\newblock Constraint handing in evvicient global optimization.
\newblock In \emph{Genetic and Evolutionary Computation Conference}, pages
  673--680. ACM, 2017.

\bibitem[Balandat et~al.(2020)Balandat, Karrer, Jiang, Daulton, Letham, Wilson,
  and Bakshy]{balandat2020botorch}
M.~Balandat, B.~Karrer, D.~R. Jiang, S.~Daulton, B.~Letham, A.~G. Wilson, and
  E.~Bakshy.
\newblock {BoTorch: A Framework for Efficient Monte-Carlo Bayesian
  Optimization}.
\newblock In \emph{Advances in Neural Information Processing Systems 33}, 2020.
\newblock URL \url{http://arxiv.org/abs/1910.06403}.

\bibitem[Berkenkamp et~al.(2016)Berkenkamp, Krause, and
  Schoellig]{Berkenkamp2016BayesianOW}
F.~Berkenkamp, A.~Krause, and A.~P. Schoellig.
\newblock Bayesian optimization with safety constraints: Safe and automatic
  parameter tuning in robotics.
\newblock \emph{ArXiv}, abs/1602.04450, 2016.

\bibitem[Chen et~al.(2021)Chen, Liu, and Tang]{chen2021new}
W.~Chen, S.~Liu, and K.~Tang.
\newblock A new knowledge gradient-based method for constrained bayesian
  optimization, 2021.

\bibitem[Cinlar(2011)]{StochProc}
E.~Cinlar.
\newblock \emph{Probability and Stochastics}, volume Graduate Texts in
  Mathematics 261.
\newblock Springer, 2011.

\bibitem[Eriksson et~al.(2019)Eriksson, Pearce, Gardner, Turner, and
  Poloczek]{Eriksson2019}
D.~Eriksson, M.~Pearce, J.~Gardner, R.~D. Turner, and M.~Poloczek.
\newblock Scalable global optimization via local bayesian optimization.
\newblock In H.~Wallach, H.~Larochelle, A.~Beygelzimer, F.~d\textquotesingle
  Alch\'{e}-Buc, E.~Fox, and R.~Garnett, editors, \emph{Advances in Neural
  Information Processing Systems 32}, pages 5496--5507. Curran Associates,
  Inc., 2019.
\newblock URL
  \url{http://papers.nips.cc/paper/8788-scalable-global-optimization-via-local-bayesian-optimization.pdf}.

\bibitem[Forrester et~al.(2008)Forrester, Sobester, and Keane]{Forrester08}
A.~I.~J. Forrester, A.~Sobester, and A.~J. Keane.
\newblock \emph{Engineering Design via Surrogate Modelling}.
\newblock 2008.

\bibitem[Frazier(2018)]{frazier2018tutorial}
P.~I. Frazier.
\newblock A tutorial on bayesian optimization, 2018.

\bibitem[Gardner et~al.(2014)Gardner, Kusner, Xu, Weinberger, and
  Cunningham]{Gardner2014}
J.~Gardner, M.~Kusner, E.~Xu, K.~Weinberger, and J.~Cunningham.
\newblock Bayesian optimization with inequality constraints.
\newblock volume~3, 06 2014.

\bibitem[Gramacy et~al.(2016)Gramacy, Gray, Digabel, Lee, Ranjan, Wells, and
  Wild]{Gramacy2016}
R.~B. Gramacy, G.~A. Gray, S.~L. Digabel, H.~K.~H. Lee, P.~Ranjan, G.~Wells,
  and S.~M. Wild.
\newblock Modeling an augmented lagrangian for blackbox constrained
  optimization.
\newblock \emph{Technometrics}, 58\penalty0 (1):\penalty0 1--11, 2016.
\newblock \doi{10.1080/00401706.2015.1014065}.
\newblock URL \url{https://doi.org/10.1080/00401706.2015.1014065}.

\bibitem[Henrnandez-Lobato et~al.(2014)Henrnandez-Lobato, Hoffman, and
  Ghahramani]{Hernandez-Lobato2014}
J.~M. Henrnandez-Lobato, M.~W. Hoffman, and Z.~Ghahramani.
\newblock Predictive entropy search for efficient global optimization of
  black-box functions.
\newblock In \emph{Proceedings of the 27th International Conference on Neural
  Information Processing Systems - Volume 1}, NIPS’14, page 918–926,
  Cambridge, MA, USA, 2014. MIT Press.

\bibitem[Hernandez-Lobato et~al.(2016)Hernandez-Lobato, Gelbart, Adams,
  Hoffman, and Ghahramani]{Hernandez-Lobato2016}
J.~M. Hernandez-Lobato, M.~A. Gelbart, R.~P. Adams, M.~W. Hoffman, and
  Z.~Ghahramani.
\newblock A general framework for constrained bayesian optimization using
  information-based search.
\newblock \emph{J. Mach. Learn. Res.}, 17\penalty0 (1):\penalty0 5549–5601,
  Jan. 2016.
\newblock ISSN 1532-4435.

\bibitem[Hern\'{a}ndez-Lobato et~al.(2016)Hern\'{a}ndez-Lobato, Gelbart, Adams,
  Hoffman, and Ghahramani]{Hernandez2016}
J.~M. Hern\'{a}ndez-Lobato, M.~A. Gelbart, R.~P. Adams, M.~W. Hoffman, and
  Z.~Ghahramani.
\newblock A general framework for constrained bayesian optimization using
  information-based search.
\newblock \emph{J. Mach. Learn. Res.}, 17\penalty0 (1):\penalty0 5549–5601,
  Jan. 2016.
\newblock ISSN 1532-4435.

\bibitem[{Jones} et~al.(1998){Jones}, {Schonlau}, and {Welch}]{Jones1998}
D.~{Jones}, M.~{Schonlau}, and W.~{Welch}.
\newblock Efficient global optimization of expensive black-box functions.
\newblock \emph{Journal of Global Optimization}, 13:\penalty0 455--492, Jan
  1998.
\newblock ISSN 0018-9219.
\newblock \doi{https://doi.org/10.1023/A:1008306431147}.

\bibitem[Lam and Willcox(2017)]{Lam2017}
R.~R. Lam and K.~E. Willcox.
\newblock Lookahead {B}ayesian optimization with inequality constraints.
\newblock In \emph{Proceedings of the 31st International Conference on Neural
  Information Processing Systems}, NIPS’17, page 1888–1898, Red Hook, NY,
  USA, 2017. Curran Associates Inc.
\newblock ISBN 9781510860964.

\bibitem[Letham et~al.(2017)Letham, Karrer, Ottoni, and Bakshy]{Letham2017}
B.~Letham, B.~Karrer, G.~Ottoni, and E.~Bakshy.
\newblock Constrained bayesian optimization with noisy experiments.
\newblock \emph{Bayesian Analysis}, 14, 06 2017.
\newblock \doi{10.1214/18-BA1110}.

\bibitem[Pearce et~al.(2020)Pearce, Klaise, and Groves]{pearce2020practical}
M.~Pearce, J.~Klaise, and M.~Groves.
\newblock Practical {B}ayesian optimization of objectives with conditioning
  variables, 2020.

\bibitem[Picheny(2014)]{picheny2014}
V.~Picheny.
\newblock A stepwise uncertainty reduction approach to constrained global
  optimization.
\newblock In S.~Kaski and J.~Corander, editors, \emph{Proceedings of the
  Seventeenth International Conference on Artificial Intelligence and
  Statistics}, volume~33 of \emph{Proceedings of Machine Learning Research},
  pages 787--795, Reykjavik, Iceland, 22--25 Apr 2014. PMLR.
\newblock URL \url{http://proceedings.mlr.press/v33/picheny14.html}.

\bibitem[Picheny et~al.(2016)Picheny, Gramacy, Wild, and
  Le~Digabel]{Picheny2016}
V.~Picheny, R.~B. Gramacy, S.~Wild, and S.~Le~Digabel.
\newblock Bayesian optimization under mixed constraints with a slack-variable
  augmented lagrangian.
\newblock In D.~D. Lee, M.~Sugiyama, U.~V. Luxburg, I.~Guyon, and R.~Garnett,
  editors, \emph{Advances in Neural Information Processing Systems 29}, pages
  1435--1443. Curran Associates, Inc., 2016.
\newblock URL
  \url{http://papers.nips.cc/paper/6439-bayesian-optimization-under-mixed-constraints-with-a-slack-variable-augmented-lagrangian.pdf}.

\bibitem[Poloczek et~al.(2017{\natexlab{a}})Poloczek, Wang, and
  Frazier]{NIPS2017_df1f1d20}
M.~Poloczek, J.~Wang, and P.~Frazier.
\newblock Multi-information source optimization.
\newblock In I.~Guyon, U.~V. Luxburg, S.~Bengio, H.~Wallach, R.~Fergus,
  S.~Vishwanathan, and R.~Garnett, editors, \emph{Advances in Neural
  Information Processing Systems}, volume~30. Curran Associates, Inc.,
  2017{\natexlab{a}}.
\newblock URL
  \url{https://proceedings.neurips.cc/paper/2017/file/df1f1d20ee86704251795841e6a9405a-Paper.pdf}.

\bibitem[Poloczek et~al.(2017{\natexlab{b}})Poloczek, Wang, and
  Frazier]{misopaper}
M.~Poloczek, J.~Wang, and P.~Frazier.
\newblock Multi-information source optimization.
\newblock In I.~Guyon, U.~V. Luxburg, S.~Bengio, H.~Wallach, R.~Fergus,
  S.~Vishwanathan, and R.~Garnett, editors, \emph{Advances in Neural
  Information Processing Systems}, volume~30. Curran Associates, Inc.,
  2017{\natexlab{b}}.
\newblock URL
  \url{https://proceedings.neurips.cc/paper/2017/file/df1f1d20ee86704251795841e6a9405a-Paper.pdf}.

\bibitem[Rasmussen and Williams(2006)]{Rasmussen06}
C.~E. Rasmussen and C.~K.~I. Williams.
\newblock \emph{Gaussian Processes for Machine Learning}.
\newblock MIT Press, 2006.

\bibitem[Sasena(2002)]{Sasenaphdthesis}
M.~Sasena.
\newblock \emph{Flexibility and Efficiency Enhancements For Constrained Global
  Design Optimization with Kriging Approximations}.
\newblock PhD thesis, 08 2002.

\bibitem[Schonlau et~al.(1998)Schonlau, Welch, and Jones]{schonlau1998}
M.~Schonlau, W.~Welch, and D.~Jones.
\newblock \emph{Global versus local search in constrained optimization of
  computer models}, volume~34, pages 11--25.
\newblock 01 1998.
\newblock \doi{10.1214/lnms/1215456182}.

\bibitem[Scott et~al.(2011)Scott, Frazier, and Powell]{Frazier211}
W.~Scott, P.~Frazier, and W.~Powell.
\newblock The correlated knowledge gradient for simulation optimization of
  continuous parameters using gaussian process regression.
\newblock \emph{SIAM Journal on Optimization}, 21\penalty0 (3):\penalty0
  996--1026, 2011.
\newblock \doi{10.1137/100801275}.
\newblock URL \url{https://doi.org/10.1137/100801275}.

\bibitem[{Shahriari} et~al.(2016){Shahriari}, {Swersky}, {Wang}, {Adams}, and
  {de Freitas}]{deFreitas2016}
B.~{Shahriari}, K.~{Swersky}, Z.~{Wang}, R.~P. {Adams}, and N.~{de Freitas}.
\newblock Taking the human out of the loop: A review of bayesian optimization.
\newblock \emph{Proceedings of the IEEE}, 104\penalty0 (1):\penalty0 148--175,
  2016.

\bibitem[Wu and Frazier(2017)]{wu2017discretizationfree}
J.~Wu and P.~I. Frazier.
\newblock Discretization-free knowledge gradient methods for {B}ayesian
  optimization, 2017.

\end{thebibliography}

% \input{100-CheckList/checklist}

% \newpage
\appendix

\section{Synthetic Test Functions}\label{App:synthetic_tests}
The following subsections describe the synthetic test functions used for the empirical comparison \citep{Sasenaphdthesis}. 

\subsection{Mystery Function}

%Consider the \f,
%\begin{gather*}
\begin{flalign*}
&\min f(x) = 2 + 0.01 (x_{2} - x_{1} ^ {2}) ^{2} + (1 - x_{1}) ^ {2} + 2 * (2 - x_{2}) ^{2} + 7  \text{sin}(0.5 x_{1}) \text{sin}(0.7  x_{1}  x_{2}) & \\
&\mbox{subject to}&\\
  &  -\text{sin}(x_{1} - x_{2} - \frac{\pi}{8}) \leq 0&\\
& x_{i} \in [0, 5], \forall i =1, 2&\\
\end{flalign*}
%\end{gather*}

%\begin{eqnarray*}
%\min f(x) &=& 2 + 0.01 (x_{2} - x_{1} ^ {2}) ^{2} + (1 - x_{1}) ^ {2} + 2 * (2 - x_{2}) ^{2} + 7  \text{sin}(0.5 x_{1}) \text{sin}(0.7  x_{1}  x_{2})  \\
%\mbox{subject to}\\
%&&-\text{sin}(x_{1} - x_{2} - \frac{\pi}{8}) \leq 0\\
%&&x_{i} \in [0, 5], \quad \forall i =1, 2\\
%\end{eqnarray*}

% \begin{figure}[h]
% 	\centering
%     	\begin{tabular}{c}
%     	Mistery\\
% 		\includegraphics[height=4.0cm]{Pics/mistery_function.jpg}\\
%     	\end{tabular}
%         \caption{Feasible (color map) and unfeasible (grey) regions of the Mistery function.The location of the optimum is highlighted by a green cross.}
% \end{figure}

\subsection{New Branin Function}

%Consider the \f,
\begin{flalign*}
&\min f(x) = -(x_{1} - 10)^{2} -(x_{2} - 15)^{2} &\\
&\mbox{subject to}&\\
&\bigg(x_{2} - \frac{5.1}{4  \pi^{2}}x_{1}^{2} + \frac{5}{\pi}x_{1} - 6\bigg)^{2} + 10 \bigg(1 - \frac{1}{8  \pi}\bigg)\text{cos}(x_{1}) + 5 \leq 0\\
&x_{1} \in [-5, 10]&\\
&x_{2} \in [0, 15]&\\
\end{flalign*}

% \begin{figure}[h]
% 	\centering
%     	\begin{tabular}{c}
%     	New Branin\\
    	
%     	\includegraphics[height=4.0cm]{Pics/branin_function.jpg}\\
%     	\end{tabular}
%         \caption{Feasible (color map) and unfeasible (grey) regions of the New Branin function. The location of the optimum is highlighted by a green cross.}
% \end{figure}

\subsection{Test Function 2}
\begin{flalign*}
&\min f(x) =  -(x_{1} - 1)^{2} -(x_{2}  - 0.5 )^{2}&\\
&\mbox{subject to}&\\
&(x_{1} - 3)^{2} + (x_{2} + 2)^{2} - 12 \leq 0&\\
&10x_{1} + x_{2} -7  \leq 0&\\
&(x_{1} - 0.5)^{2} + (x_{2} - 0.5)^{2} - 0.2 \leq 0&\\
&x_i\in[0,1]\forall i=1,2&\\
\end{flalign*}

% \begin{figure}[h]
% 	\centering
%     	\begin{tabular}{c}
%     	Test Function 2\\
    	
%     	\includegraphics[height=4.0cm]{Pics/test_function.jpg}\\
%     	\end{tabular}
%         \caption{Feasible (color map) and unfeasible (grey) regions of the Test function function.The location of the optimum is highlighted by a green cross.}

% \end{figure}

\section{MNIST Hyperparameter Experiment}\label{MNIST Hyperparameter Experiment}
% \begin{verbatim}

\textbf{Design Space:}
\begin{itemize}[leftmargin=*]
    \item \verb|learning_rate| $\in  [0.0001, 0.01]$, log scaled.
    \item \verb|beta_1| $\in  [0.7, 0.99]$, log scale.
    \item \verb|beta_2| $\in  [0.9, 0.99]$, log scale.
    \item \verb|dropout_rate_1| $\in  [0, 0.8]$, linear scale.
    \item \verb|dropout_rate_2| $\in  [0, 0.8]$, linear scale.
    \item \verb|dropout_rate_3| $\in  [0, 0.8]$, linear scale.
    \item \verb|n_neurons_1| $\in  [3, 12]$, no scaling.
    \item \verb|n_neurons_2| $\in  [3, 12]$, no scaling.
    \item \verb|n_neurons_3| $\in  [3, 12]$, no scaling.
\end{itemize}

\textbf{Neural Network Architecture:}

\begin{verbatim}

model = Sequential()

model = Dense(units = int(power(2,n_neurons_1)), input_shape=(784,))
model = Dropout(dropout_rate_1)
model = activation('relu')

model = Dense(units = int(power(2, n_neurons_2)))
model = Dropout(dropout_rate_2)
model = activation('relu')

model = Dense(units = int(power(2, n_neurons_3)))
model = Dropout(dropout_rate_3)
model = activation('relu')

model = Dense(units = 10)
model = activation('softmax')

\end{verbatim}

\textbf{Optimiser and Compilation:}

\begin{verbatim}
adam = Adam(learning_rate=learning_rate, 
            beta_1=beta_1, 
            beta_2=beta_2)


model.compile(loss='categorical_crossentropy',
              optimizer=adam,
              metrics=['accuracy'])
\end{verbatim}

\section{Related Algorithms}\label{App:algorithms_background}

\subsection{Constrained Expected Improvement (cEI)}

\cite{schonlau1998} extends EI to deterministic constrained problems by multiplying it with the probability of feasibility in the acquisition function:

$$
\text{cEI}(x|f^{*}) = \text{EI}(x|f^{*})\text{PF}^{n}(x)
$$

where $\text{PF}^{n}(x)$ is the probability of feasibility of $x$ and $\text{EI}(x|f^{*})$ is the expected improvement over the best feasible sampled observation, $f^{*}$, i.e., 

$$
\text{EI}(x|f^{*}) = \mathbb{E}[\max(y-f^{*},0)].
$$

The posterior Gaussian distribution with mean $\mu^{n}_{y}(x)$ and variance $k_{y}^{n}(x,x)$ offers a closed form solution to $\text{EI}$ where the terms only depend on Gaussian densities and cumulative distributions,

$$
\text{EI}(x|f^{*}) = (\mu^{n}_{y}(x) - f^{*}) \Phi (z) + k_{y}^{n}(x,x) \phi (z)\text{, where }z=\frac{\mu^{n}_{y}(x) - f^{*}}{k^{n}(x,x)}
$$

\subsection{Noisy Expected Improvement (NEI)}

\cite{Letham2017} further extend $\text{cEI}$ to include a noisy \f and noisy constraints. If we denote the objective and constraint values at observed \x locations as $\tilde{\mathbf{f}}^{n} = [f^{n}(x_{1}),\dots, f^{n}(x_{n})]$ and $\tilde{\mathbf{c}}^{n} = [\mathbf{c}^{n}(x_{1}),\dots, \mathbf{c}^{n}(x_{n})]$, then samples from the GP posteriors for the noiseless values of the objective and constraints at the observed points provide different estimations of $f^{*}$. Finally, NEI may be found by marginalising the possible $f^{*}$ at the sampled locations as,

$$
\text{NEI}(x) = \int_{y^{n}, c^{n}} \text{cEI}(x|\tilde{\mathbf{f}}^{n}, \tilde{\mathbf{c}}^{n}) p(\tilde{\mathbf{f}}^{n}| \Fn) p(\tilde{\mathbf{c}}^{n}| \Cn) \text{d}\tilde{\mathbf{f}}^{n} \text{d}\tilde{\mathbf{c}}^{n}
$$

Actual computations of the expectation requires a Monte-Carlo approximation, which can be computed efficiently using quasi-Monte Carlo integration.

\subsection{Knowledge Gradient KG}
\cite{Frazier211} propose the knowledge-gradient with correlated beliefs
(KG) acquisition function, which
measures the \x that attains the maximum of,

\begin{align}
\begin{split}
\text{KG}(x) = \mathbb{E}[\max_{x \in \mathbb{X}}\big\{\mu_{y}^{n+1}(x) \big\} - \max_{x \in \mathbb{X}}\big\{\mu_{y}^{n}(x) \big\}| x^{n+1} = x] 
\end{split}\label{equ:standard_KG_formulation}
\end{align}

Different approaches have been developed to solve Equ. \ref{equ:standard_KG_formulation}. \cite{Frazier211} propose discretising the \X and solve a series of linear problems. However, increasing the number of dimensions requires more discretisation points and thus renders this approach computationally expensive. A more recent approach involves Monte-Carlo sampling \citep{wu2017discretizationfree} where the \X is not discretised. Using Monte-Carlo samples improves the scalability of the algorithm but at the same time increases the computational  complexity. \citep{pearce2020practical} consider a hybrid between between both approaches that consists of obtaining high value points from the predictive posterior GP mean that would serve as a discretisation. Combining both approaches allows to leverage  the scalability of the Monte-Carlo based acquisition function and the computational performance of discretising the \X.

\subsection{Thompson Sampling with constraints (TS)}

\cite{Eriksson2019} extend Thompson sampling to constraints. Let $x_{1}, \dots, x_{r}$ be candidate points. Then
a realization is taken at the candidate points location $(\hat{f}(x_{i}), \hat{c}_{1}(x_{i}), \dots , \hat{c}_{m}(x_{i}))$
for all $x_{i}$ with 1 $\leq i \leq r$ from the respective posterior
distributions. Therefore, if $\hat{F} =\{x_{i}| \hat{c}_{l}(x_{i}) \leq 0\text{ for }1 \leq l \leq m\}$ is not empty, then the next \x is selected by $\amax_{x \in \hat{F}} \hat{f}(x)$. Otherwise a
point is selected according to the minimum total violation $\sum_{l=1}^{m} \max\{\hat{c}_{l}(x),0 \}$. 

\cite{Eriksson2019} further implements a strategy for high-dimensional \X problems based on the trust region that confines samples locally and study the effect of different transformations on the objective and constraints. However, for comparison purposes, we only implement the selection criteria.

\subsection{Constrained Predicted Entropy search (PESC)}

\cite{Hernandez-Lobato2016} seek to maximise the information about the optimal location $x^{*}$, the constrained global minimum by the acquisition function as the mutual information between y and x? given the collected data, as,

$$
\text{PESC}(x) = \text{H}(y|\Fn, \Cn) - \mathbb{E}_{x^{*}}[\text{H}[y| \Fn, \Cn, x, x^{*}]]
$$

The first term on the right-hand side of is computed as the entropy of a product of independent Gaussians. However, the second term in the right-hand side of has to be approximated.
The expectation is approximated by averaging over samples of $\hat{x}^{*} \sim \text{p}(x^{*}|\Fn, \Cn)$. To
sample $x^{*}$, first, samples from $f$ and $c_{1}, \dots , c_{K}$
are drawn from their GP posteriors. Then,  a constrained optimisation problem is solved using the sampled functions to yield a sample $\hat{x}^{*}$. 

\subsection{Penalised Knowledge Gradient (pKG)}

\cite{chen2021new}  extend KG to constrained problems by penalising any new sample by the probability of feasibility, i.e.,

\begin{align}
\begin{split}
\text{pKG}(x) = \mathbb{E}\left[\max_{x \in \mathbb{X}}\big\{\mu_{y}^{n+1}(x) \big\} - \max_{x \in \mathbb{X}}\big\{\mu_{y}^{n}(x) \big\}| x^{n+1} = x\right]  \text{PF}^{n}(x^{n+1} = x).
\end{split}\label{equ: cKG_outer_penalisation}
\end{align}

This acquisition function immediately discourages exploration in regions of low probability of feasibility and the one-step-lookahead is only on the unpenalised objective function. In their work, they extend their formulation to batches and propose a discretisation-free Monte-Carlo approach based on  \cite{wu2017discretizationfree}.

\section{Implementation Details of cKG}\label{App:Implementation Details}

Implementing $\acq$ first requires to generate $Z_{y}$ and $\mathbf{Z}_{\mathbf{c}}$ for a candidate sample $x$. This may be done by randomly generating values from a standard normal distribution, or taking Quasi-Monte samples which provides more sparse samples and faster convergence properties \citep{Letham2017}. However, we choose to adopt the method proposed by \cite{pearce2020practical} where they use different Gaussian quantiles for the objective $Z_{y} =\{\Phi^{-1}(0.1), \dots,  \Phi^{-1}(0.9) \}$. We further extend this method by also generating Gaussian quantiles for each constraint $k=1, \dots, K$ and produce the $n_{z}$ samples using the Cartesian product between the z-samples for $y$ and $k=1,\dots, K$. Once
a  set of $n_{z}$ samples has been produced, we may find each sample in $X_{d}$ by a L-BFGS optimiser, or any continuous deterministic optimisation algorithm. Finally, $KG_{d}$ in Alg.~\ref{alg:cKG_computation} may be computed using the algorithm described in Alg.~\ref{alg:KG_discretised} by \cite{Frazier211}.

To optimise $\acq$ we first select an initial set of candidates according to a Latin-hypercube design and compute their values. We then select the best subset according to their $\acq$ value and proceed to fine optimise each selected candidate design vector. We have noticed that discretisations, $X_{d}$, achieved by this subset of candidates do not change considerably during the fine optimisation, therefore we fix the discretisation found for each candidate and then fine optimise. A fixed discretisation allows to use a deterministic and continuous optimiser where approximate gradients may also be computed.

% This reduces the time complexity  to $O(n_{c}(n_{z}+1)\log(n_{z}+1))$ compared to the quadratic complexity due to computing the discretisation $X_{d}$ for the fine optimisation. Moreover

\begin{algorithm}
	\caption{Knowledge Gradient by discretisation. This algorithm takes as input a set of linear functions parameterised by a vector of intercepts $\mu$ and a vector $\tilde{\sigma}$
	}\label{alg:KG_discretised}

	\KwIn{$\mu$, $\tilde{\sigma}$, and best current \yc $\mu^{*}$}
	\vspace{1.2mm}
	0. $O \gets \text{order}(\tilde{\sigma})$ \hspace{30mm}\text{  (get sorting indices of increasing $\tilde{\sigma}$)}\\
    \vspace{1.2mm}
    1. $\mu \gets \mu[O]$, $\tilde{\sigma} \gets \tilde{\sigma}[O]$\hspace{21mm}\text{ (arrange elements)}\\
    \vspace{1.2mm}
    2. $I \gets [0,1]$\hspace{38mm}\text{(indices of elements in the epigraph)}\\
    \vspace{1.2mm}
    3. $\tilde{Z} \gets [-\infty, \frac{\mu_{0}-\mu_{1}}{ \tilde{\sigma}_{1}-\tilde{\sigma}_{0} }]$\hspace{25mm}\text{(z-scores of intersections on the epigraph)}\\
    \vspace{1.2mm}
	4. \textbf{for} j\textbf{ in }[2, \dots,  $n_{z}-1$] \textbf{:}\\
	\vspace{1.0mm}
    5.\hspace{10mm}$j \gets \text{last}(I)$\\
    \vspace{1.0mm}
    6.\hspace{10mm}$z \gets [-\infty, \frac{\mu_{i}-\mu_{j}}{ \tilde{\sigma}_{j}-\tilde{\sigma}_{i}}]$\\
    \vspace{1.0mm}
    7.\hspace{10mm}\textbf{if} $z < \text{last}(\tilde{Z})$:\\
    \vspace{1.0mm}
    8.\hspace{15mm}\text{Delete last element of $I$ and $\tilde{Z}$}.\\
    \vspace{1.0mm}
    9.\hspace{15mm}\text{Return to Line 5}.\\
    \vspace{1.0mm} 
    10.\hspace{8.5mm}\text{Add i to the end of $I$ and $z$ to $\tilde{Z}$}\\
    \vspace{1.0mm}
    11. $\tilde{Z} \gets [\tilde{Z}, \infty]$\\
    \vspace{1.0mm}
    12. $A \gets \phi(\tilde{Z}[1:]) - \phi(\tilde{Z}[:-1])$\\
    \vspace{1.0mm}    
    13. $B \gets \Phi(\tilde{Z}[1:]) - \Phi(\tilde{Z}[:-1])$\\
    \vspace{1.0mm}
    14. $\text{KG} \gets B^{T}\mu[I] - A^{T} \tilde{\sigma}[I] - \mu^{*}$ \\
    \vspace{1.0mm}
    15. \textbf{Return:} \text{KG}
\end{algorithm}

\section{Theoretical Results}

In this section we further develop the statements in the main paper. In Theorem \ref{thm:infinite_state_visitation} we show that in a discrete domain $X$ all \xs are sampled infinitely often. This ensures that the algorithm learns the true expected reward for all design vectors. In Theorem \ref{thm:reverese_xstar} we show how cKG will find the true optimal solution $x^*$ as well in the limit. 

To prove Theorem \ref{thm:infinite_state_visitation}, we rely on Lemma \ref{lemma:cKG_positive}, Lemma \ref{lemma:cKG_inf_sampling}, and Lemma \ref{lemma:cKG_finite_sampling_positive_cKG}. These ensure that \xs that are infinitely visited would not be further visited therefore visiting other states with positive \acqspace value.

\begin{lm}\label{lemma:cKG_positive}
Let $x \in \mathbb{X}$, then $cKG(x) \geq 0$
\end{lm}

\vspace{1 mm}
\textit{Proof:}

If we take the \xr according to $x^{n}_{r} = \amax_{x \in \mathbb{X}}{\mu_{y}^{n}(x)\text{PF}^{n}(x)}$ to compute the proposed formulation,

\begin{align*}
\begin{split}
\acq(x) &= \mathbb{E}[\max_{x' \in \mathbb{X}}\big\{\mu_{y}^{n+1}(x') \text{PF}^{n+1}(x')  \big\} - \mu_{y}^{n+1}(x^{n}_{r}) \text{PF}^{n+1}(x^{n}_{r}) | x^{n+1} = x].
\end{split}
\end{align*}

Then, it results straightforward to observe that the first term in the left-hand-side has a value greater or equal to the second term given by the inner optimisation operation.\qed

% TODO: simplify and change.

% We may express the maximum value of the posterior mean at iteration $n+1$ by conditional expectations as, 

% \begin{align*}
% \begin{split}
% \mathbb{E}_{\mathbf{c}^{n+1}}[\mathbb{E}_{y^{n+1}}[\max_{x' \in \mathbb{X}}\big\{\mu_{y}^{n+1}(x') \text{PF}^{n+1}(x') \big\}| x^{n+1}]]
% \end{split}\label{eq:cKG_n1}
% \end{align*}

% Let's only consider the terms inside the expectation over $\mathbf{c}^{n+1}$. We apply Jensen's inequality to the expectation over $y^{n+1}$ where the convexity of the max, and the mutual independence of $y^{n+1}$ and $\mathbf{c}$ results in,

% \begin{align*}
% \begin{split}
% \mathbb{E}_{y^{n+1}} \bigg[\max_{x' \in \mathbb{X}}\big\{\mu_{y}^{n+1}(x') \text{PF}^{n+1}(x')  \big\}| x^{n+1} \bigg] &\geq \max_{x' \in \mathbb{X}}\big\{\mathbb{E}_{y^{n+1}}\big[\mu_{y}^{n+1}(x')\big] \text{PF}^{n+1}(x';x^{n+1})  \big\}
% \end{split}\\
% \begin{split}
% &=\max_{x' \in \mathbb{X}}\big\{\mu_{y}^{n}(x') \text{PF}^{n+1}(x'; x^{n+1} )  \big\}  \end{split}\\
% \begin{split}
% &\geq \mu_{y}^{n}(x_{r}) \text{PF}^{n+1}(x_{r}; x^{n+1} )  \end{split}
% \end{align*}

% Where the second line is obtained by considering that $\mu_{y}^{n+1}(x') \sim N(\mu_{y}^{n}(x'); \tilde{\sigma_{y}}(x',x^{n+1}))$, and the bottom line considers using a \xr given by $\mu_{y}^{n}(x_{r}) \text{PF}^{n}(x_{r} )$. Since the argument of the expectation over $\mathbf{c}^{n+1}$ is positive in the proposed formulation, the overall expectation must also be positive.\qed

Then, Lemma \ref{lemma:cKG_inf_sampling} shows that if we infinitely sample a \x $x$ then the \acqspace value reduces to zero for that particular \xnospace.

\begin{lm}\label{lemma:cKG_inf_sampling}
Let $x \in \mathbb{X}$ and denote the number of samples taken in $x$ as $N(x)$, then $N(x) = \infty$ implies that $cKG(x) = 0$
\end{lm}

\vspace{1 mm}
\textit{Proof:}

If the \y is deterministic ($\sigma^{2}_{\epsilon}=0$) then sampling $x^{n+1}$ at any sampled \x $x$ produces $\sigma_{y}(x,x')=0$ for the following iterations (see Lemma 2 in \cite{misopaper}). Therefore, \acqspace becomes zero for those sampled locations.

When $\sigma^{2}_{\epsilon} > 0$, and given infinitely many observations at $x$, we have that $k_{y}^{\infty}(x, x) = 0$ and $k_{y}^{\infty}(x, x') = 0$ for all $x \in X$ by the positive definiteness of the kernel (see Pearce and Branke (2016) Lemma 3).  Then it easily follows that $\tilde{\sigma_{y}}(x,x') = 0$ and $\tilde{\sigma_{k}}(x,x') = 0$ for all $x \in X$ and $k=1,\dots K$. Therefore, $\text{PF}^{n+1}(x;x^{n+1},\mathbf{Z}_{c})= \text{PF}^{n}(x)$, and,

\begin{align*}
\begin{split}
\acq(x^{n+1}) &= \mathbb{E}_{\mathbf{Z}_{c}} [\mathbb{E}_{Z_{y}}[\max_{x \in X_{d}}\big\{\big [ \mu_{y}^{n}(x) + 0\cdot Z_{y} \big ] \text{PF}^{n}(x) \big\} 
\end{split}\\
\begin{split}
&- \mu_{y}^{n}(x_{r}) \text{PF}^{n}(x_{r}) | x^{n+1}, \mathbf{Z}_{c}]]
\end{split}\\[3ex]
\begin{split}
&=0
\end{split}\\
\end{align*}

where the bottom line comes from obtaining the \xr as $x_{r} = \amax \{\mu_{y}^{n}(x) \text{PF}^{n}(x) \}$.\qed

\begin{lm}\label{lemma:cKG_finite_sampling_positive_cKG}
Let $x^{n+1} \in \mathbb{X}$ be a \x for which $cKG(x^{n+1}) > 0$ then $N(x^{n+1})<\infty$
\end{lm}

\vspace{1 mm}
\textit{Proof:}

$\acq(x^{n+1}) > 0$ implies that $\tilde{\sigma_{y}}(x,x^{n+1}) >0$ and $\text{PF}^{n+1}(x;x^{n+1},\mathbf{Z}_{c})>0$ for some $x$. By Lemma~3 in \cite{NIPS2017_df1f1d20}, if $\tilde{\sigma_{y}}(x,x^{n+1}) >0$ then $k^{n}(x,x^{n+1})$ is not a constant function of $x'$. Therefore, only if $x^{n+1}$ is infinitely sampled, $k^{n}(x,x^{n+1})$ becomes a constant function and the maximiser value $x^{*}$ is perfectly known. Thus $x^{n+1}$ is not infinitely sampled.\qed\\

\begin{thm}\label{thm:infinite_state_visitation}
Let $X$ be a finite set and $B$ the budget to be sequentially allocated by cKG. Let $N(x, B)$ be the number of samples allocated to point $x$ within budget $B$. Then for all $x \in  X$ we have that $\lim_{B\rightarrow{\infty}} N(x, B) = \infty$.
\end{thm}

\vspace{1 mm}
\textit{Proof:}

Lemma \ref{lemma:cKG_positive} and Lemma \ref{lemma:cKG_finite_sampling_positive_cKG} imply that any point $x$ that is infinitely sampled will reach a lower bound. Since \acq ~recommends samples according to argmax, any \x $x$ that has been infinitely sampled will not be visited until all other \xs $x' \in X$ have $\acq(x')=0$. Therefore, $N(x, B) = \infty$ for all points. \qed

 To prove Theorem \ref{thm:reverese_xstar} we rely on Lemma \ref{existence_limit}. Complete derivation may be found in \citeauthor{StochProc}(\citeyear{StochProc}), in Proposition 2.8, however, the proposition states that any sequence of conditional expectations of an
integrable random variable under an increasing  convex function is a uniformly integrable martingale.

\begin{lm}
Let $x,x' \in X$ and $n\in \mathbb{N}$. The limits of the series $(\mu^{n}(x)$ and $(V^{n}(x,x')$ (shown below) exist.
\label{existence_limit} 
\end{lm}

\begin{align}
    \begin{split}
        \mu^{n}(x) &= \E_{n}[f(x)]
    \end{split}
\end{align}

\begin{align}
    \begin{split}
        V^{n}(x,x') &=\E_{n}[f(x,a)\cdot f(x')]
    \end{split}\\
    \begin{split}
        &= k^{n}(x,x')+ \mu^{n}(x,a)\cdot \mu^{n}(x') 
    \end{split}
\end{align}

Denote their limits by $\mu^{\infty}(x,a)$ and $V^{\infty}=((x,a),(x',a'))$ respectively. 

\begin{align}
    \begin{split}
        \lim_{n \rightarrow \infty} \mu^{n}(x)&=\mu^{\infty}(x)
    \end{split}\\
    \begin{split}
        \lim_{n \rightarrow \infty} V^{n}(x,,x')&=V^{\infty}(x,x')
    \end{split}
\end{align}

If $x'$ is sampled infinitely often, then $\lim_{n\rightarrow \infty} V^{n}(x,x') = \mu^{\infty}(x)\cdot 
\mu^{\infty}(x')$ holds almost surely.

\begin{thm}
Let's consider that the set of feasible \x $F=\{x|c_{k}(x)\leq 0 \text{ for }1 \leq k \leq K \}$ is not empty. If $\acq(x)= 0$ for all $x$ then
$\amax_{x \in X}\mu_{y}^{\infty}(x)\text{PF}^{\infty}(x)=\amax_{x \in X} f(x) \mathbb{I}_{x\in F}$.\label{thm:reverese_xstar} 
\end{thm}

\textit{Proof:}

By Proposition \ref{existence_limit}, $\lim_{n \rightarrow \infty} \tilde{k}_{j}^{n}(x,x')=\tilde{k}^{\infty}(x,x')$ a.s for all $x,x' \in X$ for $j=y,1,\dots,K$. If the posterior variance $\tilde{k}_{j}^{\infty}(x,x)=0$ for all $x \in X$ then we know the global optimiser. Therefore, let's consider the case of \xs such that $\hat{x} \in \hat{X} = \{x \in X | \tilde{k}^{\infty}(x,x)>0; c_{k}(x) \leq 0 \}$, then,

$$
\tilde{\sigma}_{y}^{\infty}(x,\hat{x})=\frac{k_{y}^{\infty}(x,\hat{x})}{\sqrt{k_{y}^{\infty}(\hat{x} ,\hat{x})+\sigma_{\epsilon}^{2}}}>0
$$

If we assume $\tilde{\sigma}_{y}^{\infty}(x_{1},\hat{x})\neq \tilde{\sigma}_{y}^{\infty}(x_{2},\hat{x})$ for $x_{1},x_{2} \in X$, then $cKG(x)$ must be strictly positive since for a value of $Z_{0} \in Z$,
$\mu^{\infty}_{y}(x_{1}) + \tilde{\sigma}_{y}^{\infty}(x_{1},\hat{x}) > \mu_{y}^{\infty}(x_{2}) + \tilde{\sigma}_{y}^{\infty}(x_{2},\hat{x}) $ for $Z>Z_{0}$ and vice versa. Therefore, $\tilde{\sigma}_{y}^{\infty}(x''',\hat{x})= \tilde{\sigma}_{y}^{\infty}(x'',\hat{x})$ must hold for any $x''', x'' \in X$ in order for $cKG(x)=0$, which results in,

$$
\frac{k^{\infty}(x''',\hat{x})}{\sqrt{k^{\infty}(\hat{x} ,\hat{x})+\sigma_{\epsilon}^{2}}} = \frac{k^{\infty}(x'',\hat{x})}{\sqrt{k^{\infty}(\hat{x} ,\hat{x})+\sigma_{\epsilon}^{2}}}
$$

Since $\sigma_{\epsilon}^{2}>0$, $k_{y}^{\infty}((x''',a),(\hat{x},\hat{a}) )-k^{\infty}((x'',a),(\hat{x},\hat{a}))=0$ and  $\tilde{\sigma}_{y}^{\infty}(x,\hat{x})$ does not change for all $x \in X$. It must follow that $\tilde{\sigma}_{y}^{\infty}(x,\hat{x})=0$. Theorem \ref{thm:infinite_state_visitation} also states that all $x$ locations will be visited which implies that $\text{PF}^{\infty}(x) = \{0,1\}$. Therefore, the optimiser is known $\amax_{x \in X}\mu_{y}^{\infty}(x)\text{PF}^{\infty}(x)=\amax_{x \in X} f(x) \mathbb{I}_{x\in F}$.\qed
% \bibliographystyle{abbrvnat}
% \setcitestyle{authoryear,open={((},close={))}}

% \bibliographystyle{abbrv}

\end{document}